\documentclass{article}

\usepackage{microtype}
\usepackage{graphicx}
\usepackage{booktabs} 

\usepackage{hyperref}

\usepackage[accepted]{icml2025}

\usepackage{amsmath}
\usepackage{amssymb}
\usepackage{mathtools}
\usepackage{amsthm}

\usepackage[capitalize,noabbrev]{cleveref}

\theoremstyle{plain}

\theoremstyle{definition}

\theoremstyle{remark}

\usepackage[textsize=tiny]{todonotes}

\icmltitlerunning{Transformer Dynamics}

\makeatletter
\gdef\@title{Transformer Dynamics: A neuroscientific approach to interpretability of large language models}
\makeatother

\begin{document}

\twocolumn[
\icmltitle{Transformer Dynamics: A neuroscientific approach to interpretability of large language models}




\vskip -0.5in

\begin{icmlauthorlist}
\icmlauthor{Jesseba Fernando}{NEU}
\icmlauthor{Grigori Guitchounts}{Ind}
\end{icmlauthorlist}

\begin{center}
{\small
$^{1}$Network Science Institute, Northeastern University\\
$^{2}$Independent\\[1ex]
\texttt{fernando.je@northeastern.edu, g.guitchounts@alumni.harvard.edu}}
\end{center}

\vskip 0.2in





\icmlkeywords{Mechanistic Interpretability, Transformers, LLMs, Autoencoder,}

\vskip 0.2in
]




\begin{abstract}
As artificial intelligence models have exploded in scale and capability, understanding of their internal mechanisms remains a critical challenge. 
Inspired by the success of dynamical systems approaches in neuroscience, here we propose a novel framework for studying computations in deep learning systems. 
We focus on the residual stream (RS) in transformer models, conceptualizing it as a dynamical system evolving across layers. 
We find that activations of individual RS units exhibit strong continuity across layers, despite the RS being a non-privileged basis. 
Activations in the RS accelerate and grow denser over layers, while individual units trace unstable periodic orbits. 
In reduced-dimensional spaces, the RS follows a curved trajectory with attractor-like dynamics in the lower layers. 
These insights bridge dynamical systems theory and mechanistic interpretability, establishing a foundation for a "neuroscience of AI" that combines theoretical rigor with large-scale data analysis to advance our understanding of modern neural networks.
\end{abstract}

\section{Introduction}
\label{intro}

Artificial intelligence models—particularly modern deep learning systems—have scaled in both size and capability at an astonishing rate \cite{bahri2024explaining}. 
Today's large language models (LLMs), vision models, and other predictive models (e.g. recommender systems, weather prediction, navigation, etc) are operating in the real world. 
Yet, despite their ubiquity, we lack a comprehensive understanding of how these models work, where understanding is meant roughly as to be able to predict how a particular change to the input or to the model would affect its output in a wide range of cases.
Our lack of understanding around such systems raises myriad concerns about their safety, fairness, and whether they might pose an existential risk to humanity \cite{amodei_concrete_2016, bengio_managing_2024}. 

Like the brain, deep neural networks are complex systems composed of billions of parameters interacting in highly non-linear ways. 
Systems with even a small number of such interacting elements can give rise to emergent behaviors that are unpredictable from a strictly bottom-up perspective \cite{lorenz_deterministic_1963}.
Consequently, it is not surprising that existing methods for investigating the workings of these models have yielded only fragmentary insights.

Current approaches in mechanistic interpretability often focus on identifying discrete circuits within neural networks—sub-networks or groups of neurons that implement particular functions \cite{heimersheim2024use, singh_what_2024}. 
While these circuit-based approaches have provided some explanatory power, they tend to mirror pitfalls of early approaches to understanding the brain. 
One historical example in neuroscience was the concept of "grandmother cells," which posited that the unit of representation in the brain may be individual neurons that encode highly specific concepts (like a single neuron firing selectively for one's grandmother) \cite{plaut_locating_2010}. 
This idea, together with sparse coding—the notion that only a small fraction of neurons are active at any one time, and that representations are distributed among populations of cells \cite{olshausen_sparse_2004}—led to a wave of work around sparse distributed representations as a way to explain how the brain encodes information. 
Yet, while artificially sparsifying model activations with sparse autoencoders (SAEs) has yielded model-specific information about which sets of activations represent monosemantic concepts, many questions about how the models compute remain unanswered \cite{wattenberg_relational_2024}.

\begin{figure*}[h]
    \centering
    \includegraphics[width=1.0\textwidth]{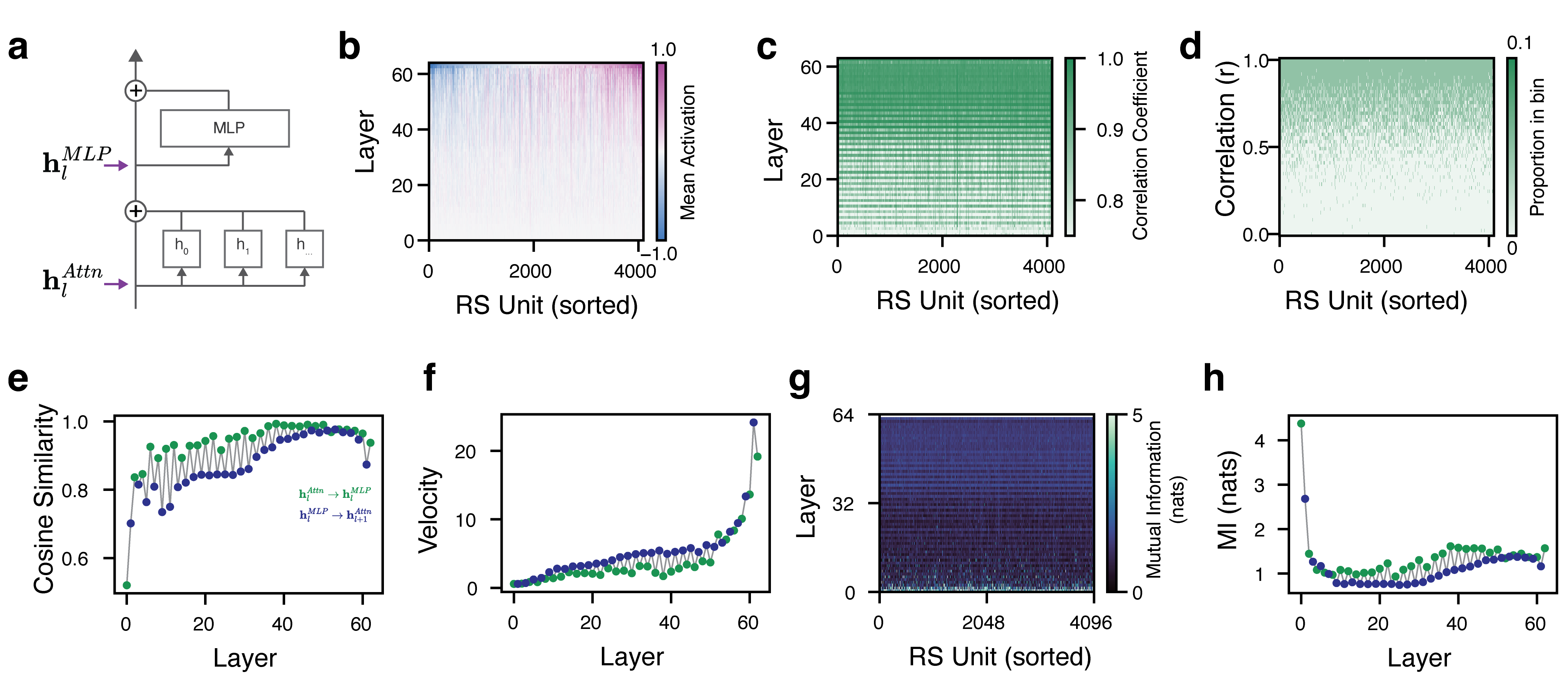}
    \vspace*{-10mm}
    \caption{ Transformer residual stream (RS) activations grow dense over the layers, are highly correlated among successive layers, and exhibit nonstationary dynamics.
    \textbf{A:} Activations of the transformer RS were captured before layernorm and the attention operation (pre-Attn) and before the MLP at each layer of Llama 3.1 8B, resulting in $\mathbf{64 \times 4096}$. 'layers' by 'units'. Activations were analyzed at the last token position for data samples from the \textsc{wikitext-2-raw-v1} dataset unless otherwise noted. 
    \textbf{B:} Mean activations across $N=1000$ samples.
    \textbf{C:} Correlations of activations for unit $u$ between layer $l$ and $l+1$ over data samples. For most units, correlations among successive layers increase over the layers.
    \textbf{D:} Histogram of correlations across layers for each unit. Despite the residual stream not having privileged basis, activations of most units are highly correlated from layer to layer.
    \textbf{E:} Cosine similarity among pairs of RS vectors $\mathbf{h}_{l}^{Attn} \rightarrow \mathbf{h}_{l}^{MLP}$ (green) and $\mathbf{h}_{l}^{MLP} \rightarrow \mathbf{h}_{l+1}^{Attn}$ (blue).
    \textbf{F:} Velocity $V$ of the RS vectors.
    \textbf{G:} Mutual information (MI) among pairs of activations for unit $u$ between layer $l$ and $l+1$ over data samples.
    \textbf{H:} MI over the layers, averaged across units in the RS.
    }
    \label{fig:fig1}
\end{figure*}

In neuroscience, a recent promising approach to interpret neural encoding of information comes from dynamical systems \cite{shenoy_cortical_2013, barack_two_2021, vyas_computation_2020}.
Treating the activity of populations of neurons as a time-evolving dynamical system has shed light on how such populations collectively implement sensory perception, compute cognitive variables, and produce behavior. 
Instead of aiming to explain the representations of single neurons—or even snapshots in time of the activities of populations—these dynamical approaches aim to understand how network-wide activity evolves over time to generate complex outputs. 
For example, in the motor system, preparatory activity appears to guide the population to an appropriate pre-movement state; and in some cases these states are attractors that are robust to noise \cite{inagaki2019discrete}. 
Dynamical approaches have yielded insights into the computations in recurrent artificial networks as well \cite{maheswaranathan_reverse_2019}. 

While transformers do not have inherent time-evolving dynamics like recurrent networks, some have examined their activations and treated them as dynamically-evolving systems \cite{geshkovski_mathematical_2024, lu_understanding_2019, hosseini_large_2023, lawson_residual_2024}. 
Specifically, the residual stream, which is updated linearly after each layer's attention and MLP operations, can be considered as a dynamic system that evolves over the layers. 
Lu et al. proposed that the transformer residual stream be considered as an ordinary differential equation (ODE) of multiple particles moving through space (i.e. across layers) and influenced by convection (external forces) and diffusion (internal forces among particles); their main contribution was proposing the Strang-Marchuk splitting scheme to replace Euler's method in approximating the ODE. 
Geshkovski et al. similarly treat the transformer as interacting particle systems, with dynamically-interacting particles (i.e. tokens) described as flows of probability measures on the unit sphere. 

This work aims to re-envision the study of mechanistic interpretability (MechInterp) through the lens of dynamical systems, inspired by this approach's success in neuroscience and driven by that field's integration of theory and large-scale data analysis. 
As such, we would like to term this new subset of MechInterp "the neuroscience of AI". Our key contributions are as follows:

\begin{enumerate}
\item We demonstrate that individual units in the residual stream maintain strong correlations across layers, revealing an unexpected continuity despite the RS not being a privileged basis.

\item We characterize the evolution of the residual stream, showing that it systematically accelerates and grows denser as information progresses through the network's layers.

\item We identify a sharp decrease in mutual information during early layers, suggesting a fundamental transformation in how the network processes information.

\item We discover that individual residual stream units trace unstable periodic orbits in phase space, indicating structured computational patterns at the unit level.

\item We show that representations in the residual stream follow self-correcting curved trajectories in reduced dimensional space, with attractor-like dynamics in the lower layers.
\end{enumerate}

\section{Methods}
\subsection{Data}

Given a corpus of text sequences from WikiText-2, we first filter the dataset $\mathcal{D}$ to include only sequences $s$ with length constraints:
\begin{equation*}
    \mathcal{D}_{\text{filtered}} = \{s \in \mathcal{D} \mid l_{\text{min}} < |s| < l_{\text{max}}\}
\end{equation*}
where $l_{\text{min}} = 100$ and $l_{\text{max}} = 500$ characters.
For each sequence $s$, we obtain its tokenized representation:
\begin{equation*}
    \mathbf{x} = [t_0, t_1, ..., t_n] = \text{tokenize}(s)
\end{equation*}
where $t_0$ is the beginning-of-sequence (BOS) token.
For the shuffled condition, we create a permuted sequence $\mathbf{x}'$ while preserving the BOS token:
\begin{equation*}
    \mathbf{x}' = [t_0, t_{\pi(1)}, t_{\pi(2)}, ..., t_{\pi(n)}]
\end{equation*}
where $\pi$ is a random permutation of indices $\{1,...,n\}$.

For each sequence (original and shuffled), we collect activations at two key points in each transformer layer $l \in \{1,...,L\}$: Pre-attention normalization ($\mathbf{h}_l^{Attn}$) and Pre-MLP normalization ($\mathbf{h}_l^{MLP}$).

These constitute the activations of the residual stream, $\mathcal{RS}$, with $\mathbf{h}_l^{Attn}$ and $\mathbf{h}_l^{MLP}$ interleaved to make up $2L$ effective `layers'.

We focused on the representation at the last token only, for each activation extracting:
\begin{equation*}
    \mathbf{h}_l \in \mathbb{R}^{B \times D}
\end{equation*}
where $B$ is the batch size and $D$ is the model dimension. 

Altogether the extracted activations corresponded to:
\begin{equation*}
    \mathcal{RS} \in \mathbb{R}^{B \times 2L \times D}
\end{equation*}

For experiments in this paper, we used Llama 3.1 8B, where $\mathrm{L=32}$ and $\mathrm{D=4096}$ (Fig\ref{fig:fig1}A).

Since activations in the RS do not correspond to individual artificial neurons, we term each dimension in $D$ a 'unit,' borrowing terminology from neuroscience to indicate the recording of the activation of a single element in the stream. 
We will thus refer to the dynamics of units in the RS as they unfold over the layers. 

\subsection{Transformer Residual Stream Activations}

Mean activations across batches for each unit and layer were calculated in the following manner:
\begin{equation*}
    \bar{\mathbf{h}}_l^{u} = \frac{1}{B}\sum_{b=1}^B \mathbf{h}_l^{i,b}
\end{equation*}

Mean activations across $N=1000$ data batches were sorted by the mean activation at the last layer: 
\begin{equation*}
    \pi = \text{argsort}(\bar{\mathbf{h}}^{2L})
\end{equation*}

\subsection{Correlations and Cosine Similarity in the Residual Stream}

For each unit $u$, we computed the Pearson correlation coefficient between its activations at different layers across all samples:
\begin{equation*}
    r_{l,l+1}^u = \frac{\text{cov}(\mathbf{h}_l^{i}, \mathbf{h}_{l+1}^{i})}{\sigma{\mathbf{h}_l^i} \sigma{\mathbf{h}_{l+1}^i}}
\end{equation*}
where $\mathbf{h}^l_{[i]} \in \mathbb{R}^B$ represents the activations of unit $i$ at layer $l$ across all samples. 

For each unit, we analyzed the distribution of correlations across all layer pairs:
\begin{equation*}
    \mathcal{C}^i = \{r_{l,m}^i : l,m \in \{1,...,2L\}, l < m\}
\end{equation*}
The distribution was binned into intervals $[0,1]$ to create a density plot.

Cosine similarity between consecutive layers was computed as:
\begin{equation*}
    \text{CS}_l = \frac{\mathbf{h}_l \cdot \mathbf{h}_{l+1}}{\|\mathbf{h}_l\| \|\mathbf{h}_{l+1}\|}
\end{equation*}

For this and other analyses, we treated the interleaved pre-attention and pre-MLP activations as layers, highlighting the two transition types: pre-attention to pre-MLP within the same layer ($\mathbf{h}_{l}^{Attn} \rightarrow \mathbf{h}_{l}^{MLP}$) and pre-MLP to pre-attention of the subsequent layer ($\mathbf{h}_{l}^{MLP} \rightarrow \mathbf{h}_{l+1}^{Attn}$).

\subsection{Velocity}

To understand how the dynamics of the residual stream change over the layers, for each layer, we calculated the magnitude of the velocity of the residual stream representation:
\begin{equation*}
    \|V_l\| = \|\mathbf{h}_{l+1} - \mathbf{h}_l\|_2
\end{equation*}

\begin{figure*}[h]
    \centering
    \includegraphics[width=1.0\textwidth]{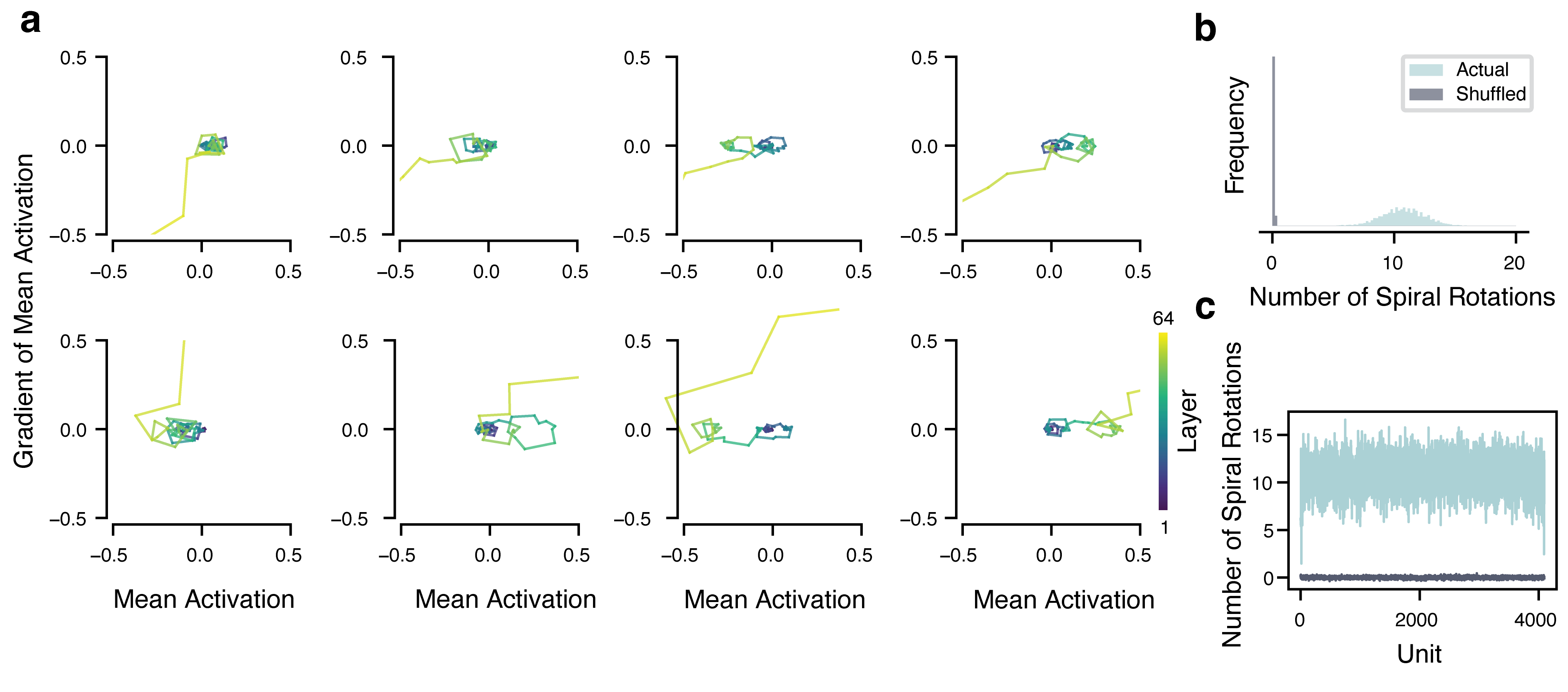}
    \vspace*{-10mm}
    \caption{
    Portraits of Individual RS Units Show Rotational Dynamics akin to Unstable Periodic Orbits. 
    \textbf{A:} Portraits of individual units in activation-gradient space, where the gradient is taken over the 64 effective sublayers. 
    \textbf{B:} Distribution of the estimated number of rotations each unit performs in this phase space compared to a control in which the layer order was shuffled 1000 times for each unit. The mean number of rotations over the layers is 10.74 for the RS units and $\sim{0}$ for the shuffle controls.
    \textbf{C:} The number of rotations for each for the 4096 units in the RS and their shuffle controls.}
    \label{fig:fig2}
\end{figure*}

\subsection{Mutual Information}

To understand how information is processed through the layers, we analyzed the mutual information (MI) of units between layers in the residual stream. 
For each pair of consecutive layers $l$ and $l+1$, we computed the mutual information using kernel density estimation:
\begin{equation*}
    \mathbf{MI}(l, l+1) = \sum p(x,y) \log \bigg(\dfrac{p(x,y)}{(p(x)p(y)}\bigg)
\end{equation*}
where $p(x,y)$ is the joint probability density of activations at layers $l$ and $l+1$, and $p(x)$ and $p(y)$ are their respective marginal densities.

We implemented this calculation using Gaussian kernel density estimation to handle the continuous nature of the activation space. 
For numerical stability, we added a small constant (1e-10) to avoid division by zero and taking logs of zero. 
The mutual information was computed in nats using the natural logarithm.

The computation was performed independently for each unit in the residual stream, using the distribution of activations derived from $N=1000$ batches, to track how different components of the representation evolved through the layers.

\subsection{Dynamics of Individual RS Units}

To examine the dynamics of individual RS units' activations across layers, we created phase portraits in 2D phase space defined by each unit's activation value and the gradient of that activation across layers.
For each unit $i$, we constructed a phase portrait where the x-axis represents the unit's activation $a^i_l$ at layer $l$, and the y-axis represents its gradient $\nabla a^i_l = \frac{d}{dl}a^i_l$.

The resulting portraits revealed rotational dynamics. 
To quantify these, the number of rotations in this 2D space was calculated by tracking the cumulative change in angle of the tangent vector along this trajectory. 
Specifically, for each unit we centered the trajectory at the origin by subtracting initial values:
$$x_l = a^i_l - a^i_0$$
$$y_l = \nabla a^i_l - \nabla a^i_0$$

We then computed tangent vectors between consecutive points:
$$\Delta x_l = x_{l+1} - x_l$$
$$\Delta y_l = y_{l+1} - y_l$$

Subsequently we calculated the angles of these tangent vectors:
\begin{equation*}
    \theta_l = \arctan2(\Delta y_l, \Delta x_l)
\end{equation*}

Following this, we computed angle changes between consecutive points, adjusting for discontinuities at $\pm\pi$:
\begin{equation*}
    \Delta \theta_l = \theta_{l+1} - \theta_l
\end{equation*}

Finally, the total number of rotations was calculated as:
\begin{equation*}
    R = \frac{1}{2\pi}\sum_{l} \Delta \theta_l
\end{equation*}

To establish statistical significance, we compared the observed number of rotations with a null distribution generated by randomly permuting the layer ordering 1000 times for each unit. 
This shuffle control preserved the distribution of activations while disrupting any systematic rotational structure.

\begin{figure*} 
    \centering
    \includegraphics[width=1.0\textwidth]{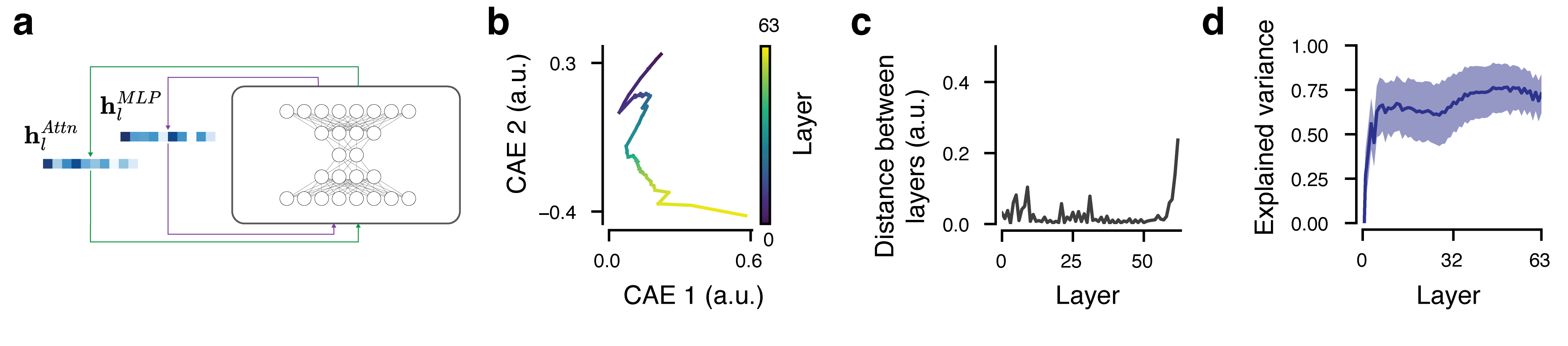}
    \vspace*{-10mm}
    \caption{Compressing Autoencoder (CAE) Shows Dynamics of the RS in Reduced Dimensional Space
    \textbf{A:} The CAE was trained to pass RS vectors at individual pre-attention and pre-MLP sublayers through a bottleneck, and reconstruct the original vector. Results showing a CAE trained with 10 layers to reduce the dimensionality at the bottleneck to 2.
    \textbf{B:} Mean trajectory across $n=1000$ test data samples.
    \textbf{C:} Distance in the reduced space between subsequent layers.
    \textbf{D:} Explained variance on the test set as a function of the layers. }
    
    \label{fig:fig3}
\end{figure*}

\subsection{Dimensionality Reduction with a Compressing Autoencoder}

To analyze the high-dimensional activation patterns in the RS, we trained an autoencoder on the RS activations and visualized the trajectories across layers in reduced dimensional space. 
The compressing autoencoder\footnote{We term this a 'compressing' autoencoder (CAE) to distinguish from Sparse Autoencoders (SAEs) in the interpretability literature.} was trained to minimize reconstruction error while learning a low-dimensional representation of activation patterns. 
The architecture consists of an encoder and decoder, each with $k$ layers where $k$ is determined by the input dimension $d_{in}=4096$ and target bottleneck dimension $d_{bottle}=2$. 
The dimensions of intermediate layers follow a geometric progression, with each layer $i$ having dimension:
\begin{equation*}
    d_i = d_{in} \cdot r^i
\end{equation*}
where $r = (d_{bottle}/d_{in})^{1/(k-1)}$ is the reduction ratio between consecutive layers.

The Wikitext dataset was used for training and evaluating the CAE, with a train set of 85k batches and test set of 15k batches, each of which contained 64 samples (i.e. an RS vector for each layer).

Each layer consists of a linear transformation followed by layer normalization and ReLU activation (except for the final encoder and decoder layers which omit these nonlinearities). The model was trained using the Adam optimizer with learning rate $\alpha=10^{-3}$ to minimize the mean squared error loss:
\begin{equation*}
    \mathcal{L}(\theta) = \frac{1}{n}\sum_{i=1}^n \|\mathbf{x}_i - f_\theta(\mathbf{x}_i)\|_2^2
\end{equation*}
where $\mathbf{x}_i$ are the input activation patterns and $f_\theta$ is the autoencoder with parameters $\theta$. 
Training proceeded for a maximum of 100 epochs with early stopping based on validation loss with a patience of 10 epochs. 
The model achieving the lowest validation loss was retained.

\subsection{PCA and Perturbation of Activation Trajectories}

To better understand the trajectories of the RS in reduced dimensional space and perform interpretable perturbations in these trajectories in reduced space, we performed Principal Component Analysis (PCA) using Singular Value Decomposition (SVD). 
For a dataset of $N$ samples with activations from $L$ layers, each of dimension $D$, we first reshape the activation tensor $\mathbf{RS} \in \mathbb{R}^{N \times 2L \times D}$ into a matrix $\mathbf{X} \in \mathbb{R}^{2NL \times D}$ by combining the sample and layer dimensions. 
Then we center the data by subtracting the mean: $\mathbf{X}_c = \mathbf{X} - \boldsymbol{\mu}$, where $$\boldsymbol{\mu} = \frac{1}{2NL}\sum_{i=1}^{2NL} \mathbf{x}_i$$

Then we compute the SVD of the centered data:
\begin{equation*}
    \mathbf{X}_c = \mathbf{U}\mathbf{\Sigma}\mathbf{V}^T
\end{equation*}
where $\mathbf{U} \in \mathbb{R}^{2NL \times D}$, $\mathbf{\Sigma_{diag}} \in \mathbb{R}^{D}$, and $\mathbf{V} \in \mathbb{R}^{D \times D}$
 
For subsequent analysis, we project the data onto the first two principal components:
\begin{equation*}
    \mathbf{Z} = \mathbf{X}_c\mathbf{V}[:,:2]
\end{equation*}
where $\mathbf{V}[:,:2]$ contains the first two right singular vectors. 
The explained variance ratio for component $k$ is computed as:
\begin{equation*}
    r_k = \frac{\sigma_k^2}{\sum_{i=1}^d \sigma_i^2}
\end{equation*}
where $\sigma_k$ is the $k$-th singular value. 

The resulting low-dimensional representation $\mathbf{Z} \in \mathbb{R}^{(2NL) \times 2}$ is then reshaped to $\mathbb{R}^{N \times 2L \times 2}$ for visualization and analysis of activation trajectories.

To investigate how perturbations in the learned low-dimensional space affect model behavior, we systematically explored the 2D PCA space by creating a uniform grid of points to which we could teleport the activations at various stages (layers) in the RS trajectory. 
Specifically, we generated $n \times n$ evenly spaced points across a range $[r_{min}, r_{max}]$ in each principal component dimension, where $n$ is the number of points per dimension (typically 10) and $r$ represents the range of perturbation magnitudes.
For each point $\mathbf{z}_i$ in this 2D grid, we projected it back to the original activation space using the inverse PCA transformation:
\begin{equation*}
    \mathbf{x}_i = \mathbf{z}_i\mathbf{V}[:2]^T + \boldsymbol{\mu}
\end{equation*}
where $\mathbf{V}[:2]$ contains the first two principal components and $\boldsymbol{\mu}$ is the mean of the original activation distribution.
We then injected these reconstructed activations into specific layers of the language model by replacing the original activations at the input to the attention layer. 
This process was repeated across multiple network layers to analyze how perturbations at different depths affect the model's internal representations.
As perturbations above the first layer we used a standard input prompt ("I'm sorry, Dave. I'm afraid I can't do that.") to record the initial trajectories. 
This input also served as a control to establish a baseline for comparison of perturbed trajectories.

\section{Results}

\subsection{Transformer residual stream (RS) activations grow dense and are highly correlated over the layers}
Our initial investigation into the activations of the RS showed that activations at the last token position for input sequences increase in magnitude over the layers (Fig. \ref{fig:fig1}B). 
Most units showed low-magnitude activations at the lowest layers, with the majority increasing in magnitude progressively over the layers. 
Sorting the mean activations across $N=1000$ data batches by the mean activation at the last layer, $\pi = \text{argsort}(\bar{\mathbf{h}}^{2L})$, revealed that activations not only grow dense as the layers progress, but that units tend to preserve their sign over the layers.

To quantify the continuity of representations between layers, we analyzed pairwise correlations between layer activations. 
We analyzed two types of transitions (Fig. \ref{fig:fig1}C): within-layer transitions ($\mathbf{h}_{l}^{Attn} \rightarrow \mathbf{h}_{l}^{MLP}$) and cross-layer transitions ($\mathbf{h}_{l}^{MLP} \rightarrow \mathbf{h}_{l+1}^{Attn}$). 

The within-layer correlations were consistently higher than the cross-layer correlations, suggesting different information processing regimes of attention and MLP operations, with the former producing smaller changes to the RS vectors. 
The correlation strength increased over layers for both transition types, with correlations starting high ($r > 0.8$) even in the earliest layers.
For each unit, the distribution of correlations over the 63 layer transitions was binned into intervals $[0,1]$ to create a density plot (Fig. \ref{fig:fig1}D), revealing that despite the RS being a nonprivileged basis, most units maintain strong correlations throughout the network.

While correlations of activations revealed that individual units exhibited ever stronger linear relationships for successive layers, we were keen to examine the changes of the RS vector as a whole. 
Cosine similarity of layerwise RS vector pairs increased as a function of layers, with within-layer ($\mathbf{h}_{l}^{Attn} \rightarrow \mathbf{h}_{l}^{MLP}$) transitions more similar than cross-layer (Fig. \ref{fig:fig1}E). 

To characterize the dynamics of representation change through layers, we computed the velocity of the RS representation. 
The velocity profile showed a distinct pattern of acceleration through the model (Fig. \ref{fig:fig1}F). 
Early layers maintain relatively constant velocities, while later layers exhibited a slight increase in velocity, with the steepest acceleration occurring in the last third of the model. 
This acceleration pattern holds for both within-layer and cross-layer transitions, though cross-layer transitions consistently showed higher velocities. 
This progressive acceleration of representational change, combined with our observations of increasing activation magnitudes and correlations, suggests that the transformer RS systematically amplifies certain representational directions in later layers. 

\begin{figure*} 
    \centering
    \includegraphics[width=1.0\textwidth]{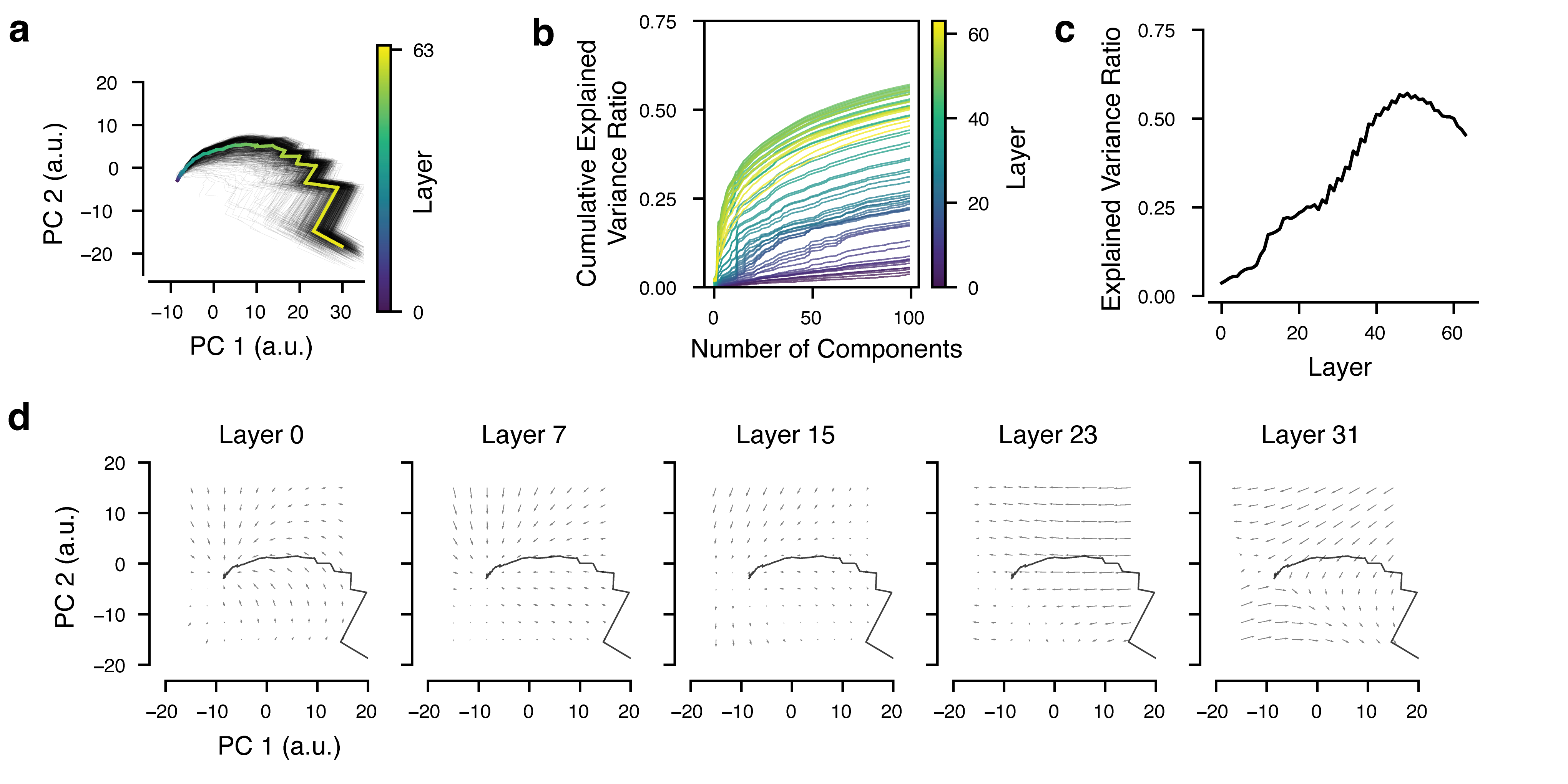}
    \vspace*{-5mm}
    \caption{Perturbation of RS trajectories Reveals Self-correcting Dynamics
    \textbf{A:} Trajectories of $n=1000$ individual (black) and mean (colored by layer) data samples in PCA space.
    \textbf{B:} Cumulative explained variance of the trajectories as a function of the number of components.
    \textbf{C:} Explained variance per layer using 100 PC components.
    \textbf{D:} Perturbation analysis in which trajectories were 'teleported' to various points, at various stages in the RS (indicated by layer number above each subplot). Gray line shows unperturbed control trajectory. Quiver arrows indicate direction and magnitude of teleported trajectories based on the successive 12 sublayers after teleportation.}
    
    \label{fig:fig4}
\end{figure*}

Analysis of mutual information (MI) for given RS units at successive layers revealed three distinct phenomena in information flow (Fig. \ref{fig:fig1}G,H). 
First, MI showed a sharp decline in early layers, with the steepest drops occurring at cross-layer transitions ($\mathbf{h}_{l}^{MLP} \rightarrow \mathbf{h}_{l+1}^{Attn}$). 
Second, the reduction in MI occured simultaneously with increasing linear correlations between layers (Fig. \ref{fig:fig1}C,D). 
Third, the MI decrease coincided with growing activation magnitudes through the layers (Fig. \ref{fig:fig1}B).

The apparent paradox between decreasing MI and increasing correlations suggests a systematic transformation of the representation space. 
While correlation captures only linear relationships, MI measures both linear and nonlinear dependencies. 
This pattern, combined with our observation of decreasing dimensionality through the layers, suggests that the model may be redistributing information across more dimensions while favoring simpler, linearly-aligned features in later layers over complex nonlinear relationships.

\subsection{RS Unit Phase Space Exhibits Rotational Dynamics}

For each unit $i$, we constructed phase portraits by plotting the unit's activation value $a^i_l$ against its gradient $\nabla a^i_l = \frac{d}{dl}a^i_l$ across layers. 
Analysis of individual RS units in activation-gradient space revealed rotational dynamics characteristic of unstable periodic orbits (Fig. \ref{fig:fig2}A). 
While most trajectories in this phase space were not particularly smooth, with abrupt changes in direction from one sublayer to the next, the overall portraits showed clear and consistent rotational patterns, with a mean of 10.74 rotations over the layers compared to approximately zero rotations in shuffle controls (Fig. \ref{fig:fig2}B). 
These rotations often spiraled out, increasing in magnitude on both axes as they progressed through the layers. 
Further, the rotations were often centered at multiple points in the phase space, not only at the origin. 

This rotational behavior was consistent across the majority of the 4096 units in the RS (Fig. \ref{fig:fig2}C), suggesting a systematic organizational principle in the network's computation. 

These observations are consistent with the data on increased velocity of the RS as a whole, and illustrate that individual RS units exhibit complex dynamics rather than simply increasing their activation magnitudes linearly.

\subsection{Reduced Dimensional Trajectories}
To analyze the high-dimensional activation patterns in the RS, we used complementary dimensionality reduction approaches: a compressing autoencoder (CAE) (Fig \ref{fig:fig3}) and PCA (Fig \ref{fig:fig4}). 
The CAE was trained to pass RS vectors through successively lower-dimensional layers, until passing through a 2-dimensional bottleneck and being reconstructed again. 
Here RS vectors at each sublayer were treated as individual samples to be learned. 
The bottleneck representation revealed structured, curving trajectories  (Fig. \ref{fig:fig3}B). 
The trajectories in this space straightened over the layers, with increasing distance between successive layers (Fig. \ref{fig:fig3}C). 
The model's reconstruction of RS vectors showed increasing explained variance, with a large jump after the early layers and a slower increase thereafter (Fig. \ref{fig:fig3}D).

Principal Component Analysis revealed that early layers distribute variance across more dimensions than later layers, with later layers requiring fewer principal components to explain the same amount of variance (Fig. \ref{fig:fig4}A,B,C).

\subsection{RS Trajectories Exhibit Attractor-like Dynamics in Lower Layers}

Visualization of RS trajectories in PCA space revealed systematic patterns in how representations evolve through the network (Fig. \ref{fig:fig4}A). 
Individual trajectories and their layer-wise means demonstrated a consistent path through this reduced space, suggesting a structured computation process, with slightly offset trajectories for the within-layer $\mathbf{h}_{l}^{Attn} \rightarrow \mathbf{h}_{l}^{MLP}$ and cross-layer $\mathbf{h}_{l}^{MLP} \rightarrow \mathbf{h}_{l+1}^{Attn}$ transitions.

To understand the stability of these trajectories, we performed perturbation analysis by "teleporting" the RS state to various points in the PCA space at different layers (Fig. \ref{fig:fig4}D). 
We hypothesized that RS progression through transformer layers might reveal attractor-like dynamics, such that moving the activations to various portions of this phase space would eventually bring them back to the mean trajectory.

The response to these perturbations varied systematically with layer depth and location of the teleported points. For instance, 

Still, the effect of the perturbations varied systematically. 
Perturbations at layer 0 (i.e. starting the trajectories at new points) resulted in highly variable dynamics, with trajectories often ending up far from the unperturbed mean.
This effect was even more drastic when interfering in the dynamics at the penultimate LLM layer, 31. 
Mid-layer perturbations (layers 7, 15, 23) exhibited a more robust recovery, with lower variance on the perturbed trajectories and lower mean squared error between the control sequence (Fig. \ref{fig:fig4}D).

This data suggests that the transformer develops stable computational channels that actively maintain desired trajectories, possibly self-correcting errors through its dynamics.

\section{Discussion}

In this work, we approached mechanistic interpretability of transformer circuits through a dynamical systems lens inspired by neuroscience. 
Treating the residual stream (RS) of the Llama 3.1 8B LLM, we found that individual RS units exhibited increasingly strong correlations from layer to layer while growing in magnitude but with a sharp drop in mutual information after the early layers. 
RS units displayed rotational structures, with dynamics reminiscent of unstable periodic orbits. 
The RS vector as a whole increased in density of activations, with more alignment of the vector at successive layers as revealed by cosine similarity.
Dimensionality reduction methods revealed low-dimensional dynamics as a whole, with trajectories that curved in low-dimensional space before straightening out and jumping progressively farther at each successive layer. 
Finally, perturbations to the RS at various layers revealed a self-correcting tendency, returning immediately close by to original spots in reduced space, akin to a pseudo-attractor.

In the broader context of mechanistic interpretability, whereas recent advances have focused on circuits \cite{singh_what_2024, kissane_interpreting_2024}  or sparse autoencoders \cite{bricken_towards_2023}, the dynamical systems approach to understanding transformers is nascent and has largely received theoretical treatment. 
Investigating the dynamics of complicated systems such as transformers could help unify theoretical insights from dynamical systems theory with large-scale data analysis and experimental manipulation.

While the present results are based on a popular open source model, Llama 3.1, preliminary analysis on another model, Gemma 2, showed similar results (data not shown). Future work beyond LLMs may examine dynamics of the residual stream in other AI architectures with a residual stream, i.e. ResNets. 
Moreover, while we presently focused on encoded sequences only at the last token position, it will be crucial to understand how whole sequences of tokens influence dynamics. 
The observed dynamics likely evolved over the course of model training, and subsequent work may investigate such dynamics over the course of learning.

Despite the high dimensionality of the residual stream (D=4096), the dynamics we observe are remarkably low-dimensional, as shown by both the autoencoder bottleneck and the interpretable structure in PCA space. 
This low-dimensional behavior may reflect the relative simplicity of our experimental task of encoding high-probability sequences of tokens from Wikitext. 
More complex tasks, like those requiring in-context learning or complex reasoning, may reveal richer dynamical structures by taking advantage of the network's representational capacity. 
This suggests future work examining how the dimensionality and structure of these dynamics scales with task complexity.

Finally, while perturbing activations by 'teleporting' them to various portions of reduced-dimensional space was revealing, it is possible this approach does not realistically capture how a model might respond to more subtle changes to its activations. Future efforts may attempt noise injection or swapping of activations from one input data sample to another.

The discovery of rotational dynamics in activation-gradient space, combined with the self-correcting properties observed in our perturbation analysis, points to an emerging organizational principle. 
The network appears to construct stable computational channels that actively maintain desired trajectories. 
This self-correction is most robust in lower layers, where cosine similarity and velocity among succeeding RS vectors are lowest, and mutual information the highest.
Finally, the strong correlations and low-dimensional flows imply that the network may perform highly distributed computations rather than localizable "grandmother cell" style encoding. 
Insights such as presented here could inform both a theoretical understanding of transformer dynamics and practical approaches to architecture design and training optimization.




\section*{Impact Statement}

This work explores the mechanistic interpretability of transformers from a dynamical systems perspective inspired by neuroscience. By bridging dynamical systems theory with transformer interpretability, we introduced a novel way to understand the behavior of LLMs. This approach follows information flows and transformations through neural networks, similar to how neural computations in the brain evolve over time. Our findings about residual stream dynamics and their self-correcting properties can inform the development of more reliable and efficient AI systems. This interdisciplinary approach unlocks a better understanding of both biological and artificial systems, and allows AI researchers to build more interpretable systems. This work contributes to the larger goal of creating AI systems that are transparent and understandable for safe and ethical deployment in society.


\bibliography{main}
\bibliographystyle{icml2025}




\end{document}